\documentclass{tmi99final}
\usepackage{gb4e}

\title{An Example-Based Approach to\\ Japanese-to-English Translation\\
of Tense, Aspect, and Modality}
\author{Masaki Murata \hspace{0.6cm} Qing Ma \hspace{0.6cm} Kiyotaka Uchimoto \hspace{0.6cm} Hitoshi Isahara}
\institute{Communications Research Laboratory, \\ Ministry of Posts and Telecommunications}
\address{588-2, Iwaoka, Nishi-ku, Kobe, 651-2401, JAPAN}
\email{\{murata,qma,uchimoto,isahara\}@crl.go.jp}
\begin{document}

\maketitle
\pagestyle{empty}

\begin{abstract}

We have developed a new method for 
Japanese-to-English translation 
of tense, aspect, and modality 
that uses an example-based method.  
In this method 
the similarity between input and example sentences is 
defined as the degree of semantic matching 
between the expressions at the ends of the sentences. 
Our method also uses the k-nearest neighbor method 
in order to exclude the effects of noise; 
for example, wrongly tagged data in the bilingual corpora. 
Experiments show that our method can 
translate tenses, aspects, and modalities more accurately 
than the top-level MT software currently available  on the market can. 
Moreover, it does not require hand-craft rules. 
\end{abstract}

\def\None#1{}

\def\Small#1{}

\def\OK#1{}

\baselineskip=1.05\baselineskip

\section{Introduction}

The translation of Japanese tenses, aspects, and modalities into English 
are some of the most difficult problems in machine translation. 
Conventional approaches to these problems 
translate Japanese tenses, aspects and modalities 
according to hand-craft rules 
that use tense and aspect information \cite{kume90e} \cite{shirai90e}. 
However, the complexity of 
Japanese tense/aspect/modality expressions makes  
it very difficult to formulate detailed rules. 
We therefore 
tried to translate Japanese tense/aspect/modality expressions 
using the example-based method, 
which was developed by Nagao \cite{NagaoEBMT}. 
We prepared bilingual corpora containing 
pairs of Japanese and English sentences 
and tried translating 
tense/aspect/modality expressions by 
using the tense/aspect/modality expression of the 
English sentence 
corresponding to the most similar Japanese sentence. 

The example-based method developed 
by Nagao in 1984 
is effective 
but has rarely been used 
since it was used 
by Sumita et al. \cite{Sumita90} 
in the translation of 
the Japanese particle {\it no}\footnote{
The Japanese particle {\it no} has many English translations: 
``of,'' ``in,'' ``at,'' ``for,'' and so on. 
Their work showed that an appropriate preposition 
can be chosen 
using the example-based method. }. 
The method we describe here is the first application of 
the example-based method 
to tense/aspect/modality translation. 
It is based on a very simple measurement of 
the similarity between an input sentence and 
an example sentence. 
Similarity is defined as the degree of 
matching between the strings 
(or the degree of semantic matching  
including the category number of the thesaurus and 
the inflectional form) 
in the expressions at the end 
of two sentences. 
Our method can also be used 
to analyze monolingual tense/aspect/modality 
if we substitute 
the corpora tagged with the correct tense/aspect/modality for 
the bilingual corpora. 

Machine translation is very difficult 
because it requires both semantic analysis and discourse analysis, 
neither of which can be done well 
by the language-processing technology 
available today. 
Since our knowledge of language analysis and generation is 
insufficient, 
we lack the fundamental knowledge needed for 
high-quality machine translation. 
But machine translation is sometimes 
accomplished well enough 
by simply replacing words, as in a puzzle game. 
We want to use even a simple technique 
if it is at all useful. 
We therefore developed 
a simple method that can translate 
tenses, aspects, and modalities better than 
the top-class MT software can but that 
does not do deep processing such as semantic analysis, 
and that does not require hand-craft rules. 

\section{Example-Based Translation of Tense, Aspect, and Modality}
\label{sec:method}

\subsection{Using a string matching at the end of sentences 
as the definition of similarity}

Murata and Nagao have already used the example-based method
to resolve the verb-phrase ellipsis 
in Japanese sentences \cite{murata0verbnlprs97}. 
In the following sentence, for example,  
the verb \citeform[I have]{arimasu} 
at the end of the sentence is omitted. 

\begin{equation}
\small
  \begin{minipage}[h]{11.5cm}
  \begin{tabular}[t]{llll}
\multicolumn{4}{l}{{\bf [Input sentence]}}\\
{\it jitsu-wa} & {\it chotto} & {\it onegaiga} & ({\it arimasu}).\\[-0.05cm]\cline{2-3}
 & & &\\[-0.35cm]
(actually)  & (a little) & (request) & (I have)\\
\multicolumn{4}{l}{
Actually, (I have) a little request.}
\end{tabular}
\end{minipage}
\end{equation}

\begin{equation}
\small
  \begin{minipage}[h]{11.5cm}
  \begin{tabular}[t]{llll}
\multicolumn{4}{l}{{\bf [Example sentence]}}\\
      &    \multicolumn{2}{c}{\bf the matching part} & {\bf the latter part}\\
{\it anou} & {\it chotto} & {\it onegaiga} & {\it arimasu}.\\[-0.05cm]\cline{2-3}
 & & &\\[-0.35cm]
(er)  & (a little) & (request) & (I have)\\
\multicolumn{4}{l}{
Er, I have a little request.}
\end{tabular}
\end{minipage}
\end{equation}

\noindent
In their method, for resolving 
this elliptical sentence, 
they search a corpus for sentences 
containing the longest string of characters matching those 
at the end of the input sentence 
(``jitsu-wa chotto onegaiga.''), 
get example sentences such as 
``anou chotto onegaiga arimasu,'' 
and judge that the verb \citeform[I have]{arimasu} 
in the latter part of the detected sentences is 
an omitted verb. 
To find an example sentence similar to the input sentence, 
we must of course first define similarity. 
How similarity is defined is critical 
because 
the result of using an example-based method depends on 
the definition of the similarity. 
Murata and Nagao 
defined the similarity as the number of characters 
in the matching part from the end of the sentence, 
a definition that is both 
simple and appropriate for this problem 
and that resolves 
the elliptical verb phrase at the end of the sentence. 

We think that 
because tense/aspect/modality expressions are 
at the ends of Japanese sentences, 
this definition of similarity can also be used 
in tense/aspect/modality translation. 
Our method searches a bilingual corpus for 
the Japanese example sentence 
containing the longest string matching that 
at the end of the input Japanese sentence, 
and it selects 
the tense/aspect/modality expression of the corresponding English sentence 
in the corpus 
as the tense/aspect/modality expression used for the English translation of 
the Japanese input sentence. 
Suppose that 
we translate the tense/aspect/modality expression of 
the following Japanese input sentence. 

\begin{equation}
\small
  \begin{minipage}[h]{11.5cm}
  \begin{tabular}[t]{llllll}
\multicolumn{6}{l}{{\bf [Input sentence]}}\\
{\it kare-wa} & {\it yuumei-ni} & {\it naritai-toiu} & {\it yashin} & {\it wo} & {\it idai-teiru}.\\[-0.05cm]\cline{5-6}
 & & &\\[-0.35cm]
(He)  & (famous) & (to become) & (an ambition) & {\sf obj} & (have)\\
\multicolumn{6}{l}{He has an ambition to become famous. }
\end{tabular}
\end{minipage}
\end{equation}

\begin{equation}
\small
  \begin{minipage}[h]{11.5cm}
  \begin{tabular}[t]{llllll}
\multicolumn{6}{l}{{\bf [Example sentence]}}\\[-0.2cm]
\multicolumn{6}{l}{\hspace*{7.5cm} \bf The matching part}\\
{\it kare-wa} & {\it hurusato-eno} & {\it hageshii} & {\it bojoh} & {\it wo} & {\it idai-teiru}.\\[-0.05cm]\cline{5-6}
 & & &\\[-0.35cm]
(He)  & (home) & (great) & (a longing) & {\sf obj} & (have)\\
\multicolumn{6}{l}{
He \underline{has} a great longing for home.}\\
\multicolumn{6}{l}{\hspace*{-1cm}\bf The corresponding part}\\
\end{tabular}
\end{minipage}
\end{equation}

\noindent
At first, we detect the example sentence containing 
the longest expression at the end of this input sentence. 
We then find that 
the above example sentence is the one containing 
the longest expression \citeform[have.]{wo idai-teiru,} 
We look at the verb of the English translation of the example, 
find that the tense/aspect/modality expression is the present tense form, 
and translate the tense/aspect/modality expression of the input sentence 
into the present tense. 
A rule-based method, in contrast, would likely 
determine the tense/aspect/modality expression to be the progressive form, 
since this sentence has 
a Japanese tense/aspect/modality expression {\it teiru}, 
which often means progression\footnote{
The Japanese tense/aspect/modality expression {\it teiru} 
often means progression as in the following sentence. 
\begin{equation}
\scriptsize
  \begin{minipage}[h]{11.5cm}
  \begin{tabular}[t]{llllll}
{\it kare-wa} & {\it sentou-no} & {\it sousha-ni} & {\it pittari-kuttsuite} & {\it hashit} & {\it \underline{-teiru}}\\[-0.05cm]
 & & &\\[-0.35cm]
(He)  & (front) & (runner) & (at the heels of) & (run) & (-ing)\\
\multicolumn{6}{l}{He is runn\underline{ing} at the heels of the front runner. }
\end{tabular}
\end{minipage}
\end{equation}}.
Our example-based method, 
however, 
can correctly judge that the tense/aspect/modality expression 
of the input sentence (3) is the present tense form.  

This similarity based on matching the strings at the end of 
sentences is 
simpler and more tractable 
than the similarity used in the 
translation of {\sf Noun X} {\it no} {\sf Noun Y}, 
to which the example-based method was first applied. 
In the problem of {\sf Noun X} {\it no} {\sf Noun Y}, 
there are some cases 
when {\sf Noun X} is more important than {\sf Noun Y} 
and some cases 
when {\sf Noun Y} is more important than {\sf Noun X}. 
We therefore need to appropriately weight {\sf Noun X} and {\sf Noun Y}, 
so the similarity is very complicated. 
But if we measure similarity by matching the strings at the end of 
sentences, we have only to check the string in order
from the end of the sentence. 

\subsection{Two measures for matching strings at the end of sentences}

In recent years, 
the technologies on natural language processing have developed 
and various morphological analyzers are open to the public. 
In our analysis, 
we check the degree of matching of strings at the end of a sentence 
in order to detect an example similar to an input sentence. 
At this time, 
we check the matching after 
recognizing words by morphological analysis. 
And we also check the similarity between words 
by using the semantic distance between words 
in the thesaurus rather than by matching strings of words. 
For checking the match at the end of a sentence, 
we therefore use the method using the result of language analysis 
in addition to the method using only strings. 
These two methods are explained below: 

\begin{itemize}
\item 
  {\bf Method 1} \underline{Using simple strings}

  This method is the one mentioned in the previous section. 
  It checks the degree of string matching from the end 
  of a sentence 
  and uses the length of the matching string as the similarity. 

\item 
  {\bf Method 2} \underline{Use of the result of language analysis}

  This method performs high-quality matching 
  by using a morphological analyzer and 
  a thesaurus. 
  At first, 
  we detect morphologies by using 
  a morphological analyzer \cite{JUMAN3.5e}. 
  Next, we give each morphology 
  a category number representing that morphology 
  in a Japanese word thesaurus \cite{NLRI64ae}. 
  When the morphology is an inflectional word, 
  we also give it 
  the inflectional form (e.g., the past tense) that is obtained from 
  the output of the morphological analyzer. 

\begin{table}[t]
\footnotesize
\caption{Information obtained from language analysis}
    \label{tab:kaisekijoho}
  \begin{center}
  \begin{tabular}[h]{lll|l|ll}\hline
    \multicolumn{3}{c|}{Morphology} & \multicolumn{1}{c|}{Category number} & \multicolumn{2}{c}{Inflectional form}\\\hline
Èà  &  {\it kare} & (He) & 1200003012 &&\\
¤Ï   & {\it wa} & {\sf topic} & 1195038023 &&\\
Ìî˾ &   {\it yabou} & (ambition) & 1304207024 &&\\
¤ò   & {\it wo} &{\sf obj} & &&\\
¤¤¤À¤¤¤Æ &   {\it idaite} & (have) & 2153417012 & ¥¿·ÏÏ¢ÍѥƷÁ &({\it ta}-series predicative {\it te}-form)\\
¤¤¤ë  &  {\it iru} & (be) & 2120002012 & ´ðËÜ·Á & (the normal form)\\\hline
  \end{tabular}
\end{center}
\end{table}

For example, 
the sentence ``{\it kare wa yabou wo idaite iru.}'' 
(``He has an ambition.''), 
is represented by the information in Table \ref{tab:kaisekijoho}. 
In the table, 
the input sentences are divided 
into morphologies such as \citeform[he]{kare} and 
{\it wa} {\sf topic}, 
and each of them is given 
a category number and the inflectional form. 
In the thesaurus, 
each word has a 10-digit category number. 
This 10-digit category number indicates
seven levels of an {\sf is-a} hierarchy. 
The top five levels are expressed by 
the first five digits of the category number. 
The sixth level is expressed 
by the following two digits of the category number. 
And the last level 
is expressed by the last three digits of the category number. 

After assigning the category numbers, 
we check the degree of matching at the end 
of a sentence by using the information in Table \ref{tab:kaisekijoho}. 
At this time, 
to check the string match from the end of the sentence 
in the same as in Method 1, 
we use the following string 
combining all the information in Table \ref{tab:kaisekijoho}. 
(We do not use the last three digits of the category number.)

  \begin{quote}
    \footnotesize
    Èà03£°£°£°£²£±¡§

    ¤Ï38£°£µ£¹£±£±¡§

    Âç˾07£²£´£°£³£±¡§

    ¤ò¡§

    ¤¤¤À¤¤¤Æ17£´£³£µ£±£²¥¿·ÏÏ¢ÍѥƷÁ¡§

    ¤¤¤ë02£°£°£²£±£²´ðËÜ·Á
  \end{quote}

In this information, we reverse the category number. 
This means that 
when we check the matching from the end of a sentence, 
we check the matching from the top of the category number 
and obtain the same result we would obtain if we used 
the normal way 
to check semantic similarity in the thesaurus. 

In Method 2, we transform an input sentence into 
the above information and check the length of the matching characters 
from the end of the sentence. 
The length of the matching characters is 
treated as the similarity used in the example-based method. 
We can check, in order, the inflectional form, 
the similarity in the thesaurus,  
and the similarity of the strings of each morphology 
by checking the string match from the end in the above information. 
\end{itemize}

\subsection{Using  the k-nearest neighbor method 
for preventing the problem of noise}

The k-nearest neighbor method 
contains the example-based method 
\cite{Fukunaga}. 
Instead of using the one-nearest example, 
this method uses 
the result obtained from the ``voting'' of the k nearest examples. 
The decision obtained by 
using only one example is unreliable 
since that example may be a noise. 
The decision using k examples 
makes a stable analysis possible 
even when the data include a little noise. 

In the work reported here, 
we used 1, 3, 5, 7, and 9 as k. 
When one of the k-nearest examples 
has the same highest similarity as other examples, 
we should use all of them regardless of the value of k. 
In this work, however, 
we limited the number of examples to 10 
in order to simplify the processing. 
When different tense/aspect/modality 
expressions had the same number of votes, 
the expression selected was that of the example obtained first. 

\begin{table*}[t]
  \scriptsize
\caption{Example of tense/aspect/modality translation}
\label{tab:kaisekirei}
\vspace*{-0.3cm}
  \begin{center}
  \begin{tabular}[h]{r|r|l|l|l}\hline
\multicolumn{2}{l|}{} & ¡¡Japanese      & \multicolumn{1}{|c|}{Category} &  ¡¡English\\\hline
\multicolumn{2}{r|}{Input} &Èà¤Ï»ä¤ÎÃΤê¹ç¤¤¤À & Present & I am acquainted with him. \\\hline
No. & Sim. & \multicolumn{2}{|l}{ Example sentence } & \multicolumn{1}{l}{  } \\\hline
1 & 25 & Èà¤È¤ÏĹǯ\underline{¤ÎÃΤê¹ç¤¤¤À} & Present perfect & I have known him for a long time.\\
2 & 24 & ¤Õ¤¿¤ê¤ÏŤ¤´Ö\underline{¤ÎÃΤê¹ç¤¤¤À} & Present & The two are acquaintances of long standing. \\
3 & 11 & Èà¤È¤Ï£±£°Ç¯Í¾¤Î\underline{´é¸«ÃΤê¤À} & Present perfect & I have known him for over ten years. \\
4 & 11 & Èà¤é¤Ï¿ǯ¤Î\underline{ÃθʤÀ} & Present & They are friends of many years' standing. \\
5 & 10 & Èà¤Ï¤³¤Î¥¯¥é¥Ö¤Î\underline{²¸¿Í¤À} & Present & He is a benefactor of this club. \\
6 & 10 & Èà¤Ï»ä¤ÎÌ¿¤Î\underline{²¸¿Í¤À} & Present & I owe him my life.   \\
7 & 10 & Èà¤Ï¤«¤¿¤¤\underline{¿Í¤À} & Present & He is reliable. \\
8 & 10 & Èà¤Ï¤À¤ì¤Ë¤â¿ÍÅö¤¿¤ê¤Î¤¤¤¤\underline{¿Í¤À} & Present & He is affable to everybody. \\
9 & 10 & Èà¤ÏÉ濨¤ê¤Î¤¤¤¤\underline{¿Í¤À} & Present & He is a mild-mannered person. \\
10 & 10 & ¤Ê¤ó¤ÈÃË¿¶¤ê¤Î¤¤¤¤\underline{¿Í¤À} & Present & What a handsomelooking man he is! \\\hline
\end{tabular}
\end{center}
Sim. = Similarity\\
Category = Category of Tense/Aspect/Modality
\end{table*}

\begin{table*}[t]
\caption{Calculation of similarity by string matching from the end of the sentence}
\label{tab:ittchijokyo}
\vspace*{-0.3cm}
  \begin{center}
\small
\scriptsize
  \begin{tabular}[h]{@{ }c@{}|@{ }c@{}|@{ }r@{}r@{}r@{}r@{}r@{}r@{}r@{}r@{}r@{}r@{}r@{}r@{}r@{}r@{}r@{}r@{}r@{}r@{}r@{}r@{}r@{}r@{}r@{}r@{}r@{}r@{}r@{}r@{}r@{}r@{}r@{}r@{}r@{}r@{}r@{}r@{}r@{}r@{}r@{}r@{}r@{}r@{}r@{}r@{}r@{}r@{}r@{}r@{}r@{}r@{}r@{}r@{}r@{}r@{}r@{}r@{}}\hline
\multicolumn{2}{r|}{Input} &  Èà & 03 & £° & £° & £° & £² & £± & ¡§ & ¤Ï & 38 & £° & £µ & £¹ & £± & £± & ¡§ & »ä & 01 & £° & £° & £° & £² & £± & ¡§ & ¤Î & 07 & £° & £° & £° & £± & £± & ¡§ & ÃÎ & ¤ê & ¹ç & ¤¤ & 03 & £° & £± & £² & £² & £± & ¡§ & ¤À & ´ð & ËÜ & ·Á \\\hline
No. & Sim. &&&&&&&&&&&&&&&&&&&&&&& {25} &24 &23 &22 &21 &20 &19 &18 &17 &16 &15 &14 &13 &12 &11 &10 &9 &8 &7 &6 &5 &4 &3 &2 &1 \\\hline
1 & 25 &  38 & £° & £µ & £¹ & £± & £± & ¡§ & ¤Ï & 38 & £° & £µ & £¹ & £± & £± & ¡§ & Ĺ & ǯ & 07 & £² & £´ & £¶ & £± & {\bf £±} & {\bf ¡§} & {\bf ¤Î} & {\bf 07} & {\bf £°} & {\bf £°} & {\bf £°} & {\bf £±} & {\bf £±} & {\bf ¡§} & {\bf ÃÎ} & {\bf ¤ê} & {\bf ¹ç} & {\bf ¤¤} & {\bf 03} & {\bf £°} & {\bf £±} & {\bf £²} & {\bf £²} & {\bf £±} & {\bf ¡§} & {\bf ¤À} & {\bf ´ð} & {\bf ËÜ} & {\bf ·Á} \\
2 & 24 &  & ¤Õ & ¤¿ & 03 & £° & £µ & £¹ & £± & £± & ¡§ & ¤ê & ¡§ & ¤Ï & 38 & £° & £µ & £¹ & £± & £± & ¡§ & Ĺ & ¤¤ & ´Ö & {\bf ¡§} & {\bf ¤Î} & {\bf 07} & {\bf £°} & {\bf £°} & {\bf £°} & {\bf £±} & {\bf £±} & {\bf ¡§} & {\bf ÃÎ} & {\bf ¤ê} & {\bf ¹ç} & {\bf ¤¤} & {\bf 03} & {\bf £°} & {\bf £±} & {\bf £²} & {\bf £²} & {\bf £±} & {\bf ¡§} & {\bf ¤À} & {\bf ´ð} & {\bf ËÜ} & {\bf ·Á} \\
.. &.. &.. &&&&&&&&&&&&&&&&&&&&&&&&&&&&&&&&&&&&&&&&&&&&&&\\
4 & 11 &  Èà & ¤é & 03 & £° & £° & £° & £² & £± & ¡§ & ¤Ï & 38 & £° & £µ & £¹ & £± & £± & ¡§ & ¿ & ǯ & 07 & £° & £± & £¶ & £± & £± & ¡§ & ¤Î & 07 & £° & £° & £° & £± & £± & ¡§ & ÃÎ & ¸Ê & {\bf 03} & {\bf £°} & {\bf £±} & {\bf £²} & {\bf £²} & {\bf £±} & {\bf ¡§} & {\bf ¤À} & {\bf ´ð} & {\bf ËÜ} & {\bf ·Á}\\
.. &.. &.. &&&&&&&&&&&&&&&&&&&&&&&&&&&&&&&&&&&&&&&&&&&&&&\\\hline
\end{tabular}
\end{center}
\end{table*}

Next we examine the k-nearest method 
by using the example of tense/aspect/modality translation 
in Table \ref{tab:kaisekirei}. 
Table \ref{tab:kaisekirei} shows 
the analysis of the tense/aspect/modality expression of 
the input sentence 
``{\it kare wa watashi no shiriai da.}'' (I am acquainted with him.) 
by using Method 2. 
The calculation of 
the similarity by using Method 2 
is illustrated by the data listed in Table \ref{tab:ittchijokyo}, 
where 
the bold-faced part 
matches the input sentence. 
One Japanese character consists of two bytes. 
So in this work, the number of two-byte sequences in the matching part 
represents the similarity. 
Example 1 in Table \ref{tab:ittchijokyo}, for example, 
has a similarity of 25 
since the length of the matching part is 25 two-byte sequences. 
The results obtained from the 10 most-similar example sentences 
are listed in Table \ref{tab:kaisekirei}, 
where ``Tense/Aspect/Modality'' is that 
obtained from the tense/aspect/modality expression of 
the English sentence corresponding to 
the Japanese example sentence. 

When k = 1, 
the tense/aspect/modality expression was analyzed by using only 
the example most similar to the input sentence, Example 1, 
which has the tense/aspect/modality expression ``present perfect.'' 
So our system judged that 
the target tense/aspect/modality expression was ``present perfect,'' 
even though the correct one was 
``present.'' 
When k = 3, 
we tried to select the three most-similar example sentences 
but found that 
Examples 3 and 4 had the same similarity. 
So we used four examples, two of which voted 
``present perfect'' and 
two of which voted ``present.'' 
The incorrect tense/aspect/modality expression ``present perfect'' 
was again selected because it was obtained earlier 
in the processing. 
When k = 5, 
we tried to select the five most-similar example sentences but 
found that Examples 5 through 10 had 
the same similarity. 
So we used all ten, two of which voted for 
``present perfect'' and 
eight of which voted ``present.'' 
The correct tense/aspect/modality expression, ``present,'' 
was thus selected. 
When k = 7 or 9, 
we used all ten and 
got the correct tense/aspect/modality expression, 
``the present,'' as when k = 5. 
The system outputted an incorrect answer 
when k is 1 or 3, 
and 
outputted a correct answer when k is 5, 7 or 9. 

\section{Experiment and Discussion}

\subsection{Experiment}

We carried out the experiments on tense/aspect/modality translation 
in order to verify the method described in Section \ref{sec:method}. 
We used 
the bilingual corpus (36,617 sentences) 
in the Kodansha Japanese-English dictionary 
\cite{kodanshawaeiE} as the database of examples. 
From this corpus, we randomly selected 300 sentences 
as input sentences and compared the results 
obtained by using our method with those obtained 
by using the top-level software 
currently available  on the market. 
When we ran the software on the 300 input sentences, 
the verb parts of 11 of them could not be translated 
and the tense/aspect/modality expressions could not be obtained from them. 
We therefore eliminated these 11 sentences 
from our experiments. 

\begin{table*}[t]
\footnotesize
    \caption{Result}
    \label{tab:daibunrui}
  \begin{center}
\begin{tabular}[c]{l|r@{\% }c|r@{\% }c|r@{\% }c|r@{\% }c}\hline
 & \multicolumn{2}{|c|}{All} & \multicolumn{2}{|c|}{Present} & \multicolumn{2}{|c|}{Past} & \multicolumn{2}{|c}{Other}\\\hline
Software &  80.6 & (233/289)& 91.1 & (112/123)& 96.3 & (105/109)& 28.1 & ( 16/ 57)\\\hline
Method 1(k=1)&  76.8 & (222/289)& 89.4 & (110/123)& 89.9 & ( 98/109)& 24.6 & ( 14/ 57)\\
Method 1(k=3)&  82.0 & (237/289)& 93.5 & (115/123)& 97.2 & (106/109)& 28.1 & ( 16/ 57)\\
Method 1(k=5)&  83.0 & (240/289)& 95.1 & (117/123)& 97.2 & (106/109)& 29.8 & ( 17/ 57)\\
Method 1(k=7)&  82.4 & (238/289)& 94.3 & (116/123)& 96.3 & (105/109)& 29.8 & ( 17/ 57)\\
Method 1(k=9)&  82.4 & (238/289)& 94.3 & (116/123)& 96.3 & (105/109)& 29.8 & ( 17/ 57)\\\hline
Method 2(k=1)&  78.5 & (227/289)& 88.6 & (109/123)& 89.9 & ( 98/109)& 35.1 & ( 20/ 57)\\
Method 2(k=3)&  81.7 & (236/289)& 91.1 & (112/123)& 95.4 & (104/109)& 35.1 & ( 20/ 57)\\
Method 2(k=5)&  81.3 & (235/289)& 92.7 & (114/123)& 93.6 & (102/109)& 33.3 & ( 19/ 57)\\
Method 2(k=7)&  81.7 & (236/289)& 92.7 & (114/123)& 93.6 & (102/109)& 35.1 & ( 20/ 57)\\
Method 2(k=9)&  81.7 & (236/289)& 92.7 & (114/123)& 93.6 & (102/109)& 35.1 & ( 20/ 57)\\\hline
\end{tabular}
\end{center}
\end{table*}

We classified the tense/aspect/modality into 
the following 27 categories: 
\begin{enumerate}
\item 
  all the combinations of \{Present, Past\}, \{Progressive¡¤Not-progressive\}, 
  and \{Perfect, Not-perfect\} (8 categories),  
\item 
  imperative mood (1 category), 
\item 
  auxiliary verbs (\{Present, Past\} of ``be able to'', \{Present, Past\} of ``be going to'', can, could, have to, had to, let, may, might, must, need, ought, shall, should, will, would) (18 categories). 
\end{enumerate}
``Must'' and ``have to'' or ``can'' and ``be able to'' 
should really be grouped together, 
but since they may have different meanings, 
we defined the tense/aspect/modality 
according to the English surface expression strictly 
and handled these cases as different tenses/aspects/modalities. 
We used 
the tense/aspect/modality expression of the corresponding verb 
in the English sentence as the correct tense/aspect/modality\footnote{
  In the experiment 
  the criterion for judging 
  whether the result was correct was 
  very strict: 
  the output tense/aspect/modality must be 
  the same as the tense/aspect/modality of the English translation 
  of the input sentence 
  in our bilingual database. 
  As in 2(b) in Section \ref{sec:discussion}, 
  there are some cases when 
  English tense/aspect/modality expressions 
  that express the same tense/aspect/modality are different. 
  The real accuracy rates may be much higher than listed 
  those in Table \ref{tab:daibunrui}. }.

\begin{table*}[t]
\scriptsize
\caption{Accuracy when determining each tense/aspect/modality}
    \label{tab:shoubunrui}
  \begin{center}
\begin{tabular}[c]{l@{ }|r@{\% }|r@{ }r@{ }r@{ }r@{ }r@{ }r@{ }r@{ }r@{ }r@{ }r@{ }r@{ }r@{ }r}\hline
 & \multicolumn{1}{|c|}{All} & \multicolumn{1}{c}{Pr.} & \multicolumn{1}{c}{Past} & \multicolumn{1}{c}{\scriptsize Pr.-ing} & \multicolumn{1}{c}{\scriptsize P.-ing} & \multicolumn{1}{c}{Perf.} & \multicolumn{1}{c}{Imp.} & \multicolumn{1}{c}{can} & \multicolumn{1}{@{}c@{}}{could} & \multicolumn{1}{@{ }c@{ }}{¡¡let¡¡} & \multicolumn{1}{c}{may} & \multicolumn{1}{c}{must}& \multicolumn{1}{c}{will} & \multicolumn{1}{@{}c@{}}{would}  \\\hline
\multicolumn{1}{l|}{No.}       &   \multicolumn{1}{|r@{ }|}{289\hspace{-0.2mm}} &  123\hspace{-0.2mm} &  109\hspace{-0.2mm} &    7\hspace{-0.2mm} &    1\hspace{-0.2mm} &   15\hspace{-0.2mm} &   12\hspace{-0.2mm} &    3\hspace{-0.2mm} &    2\hspace{-0.2mm} &    1\hspace{-0.2mm} &    2\hspace{-0.2mm} &    4\hspace{-0.2mm} &    9\hspace{-0.2mm} &    1\hspace{-0.2mm}\\\hline
\multicolumn{15}{c}{Software}\\\hline
 &    81 &   91\% &   96\% &   29\% &    0\% &    0\% &   67\% &   67\% &  100\% &  100\% &    0\% &   25\% &    0\% &    0\%\\\hline
\multicolumn{15}{c}{Method 1}\\\hline
k=1&    77 &   89\% &   90\% &   14\% &    0\% &    7\% &   58\% &   33\% &   50\% &  100\% &    0\% &    0\% &   22\% &    0\%\\
k=3&    82 &   93\% &   97\% &   29\% &    0\% &    0\% &   75\% &   33\% &    0\% &  100\% &    0\% &    0\% &   33\% &    0\%\\
k=5&    83 &   95\% &   97\% &   14\% &    0\% &    0\% &   83\% &   33\% &    0\% &  100\% &    0\% &   25\% &   33\% &    0\%\\
k=7&    82 &   94\% &   96\% &   14\% &    0\% &    0\% &   83\% &   33\% &    0\% &  100\% &   50\% &   25\% &   22\% &    0\%\\
k=9&    82 &   94\% &   96\% &   14\% &    0\% &    0\% &   83\% &   33\% &    0\% &  100\% &   50\% &   25\% &   22\% &    0\%\\\hline
\multicolumn{15}{c}{Method 2}\\\hline
k=1&    79 &   89\% &   90\% &   29\% &    0\% &   13\% &   75\% &   33\% &  100\% &  100\% &    0\% &    0\% &   33\% &    0\%\\
k=3&    82 &   91\% &   95\% &   29\% &    0\% &   13\% &   83\% &    0\% &   50\% &  100\% &    0\% &   25\% &   33\% &    0\%\\
k=5&    81 &   93\% &   94\% &   29\% &    0\% &    7\% &   92\% &    0\% &    0\% &  100\% &    0\% &   25\% &   33\% &    0\%\\
k=7&    82 &   93\% &   94\% &   14\% &    0\% &    7\% &   92\% &    0\% &    0\% &  100\% &   50\% &   50\% &   33\% &    0\%\\
k=9&    82 &   93\% &   94\% &   14\% &    0\% &    7\% &   92\% &    0\% &    0\% &  100\% &   50\% &   50\% &   33\% &    0\%\\\hline
\end{tabular}
\end{center}
\end{table*}

The accuracies obtained when determining each tense/aspect/modality 
are listed in Table \ref{tab:daibunrui}. 
Only 13 of the 27 tense/aspect/modality expressions
were found in the 289 sentences, 
and 
the accuracy rates for each of them are listed 
in Table \ref{tab:shoubunrui}. 
``Pr,'' ``P.,'' ``-ing,'' ``Perf.,'' and ``Imp.'' 
respectively 
indicate 
``Present,'' ``Past,'' ``Progressive,'' ``Perfect,'' and ``Imperative.'' 

\subsection{Discussion}
\label{sec:discussion}

\begin{enumerate}
\item 
Accuracy rates

\begin{enumerate}
\item 
  Method 1 when k=5 is best (83\%) (Table \ref{tab:daibunrui}). 
  This result shows that 
  even simple string-matching 
  can yield comparatively high accuracy rates. 

\item 
  All the overall ``All'' accuracy rates obtained using 
  our methods when k $\neq$ 1 
  are higher than those obtained using the software. 

  When k = 1, our methd suffers from 
  noise and the accuracy rate is low. 
  And the results in Table \ref{tab:daibunrui} clearly 
  show that 
  the k-nearest neighbor method is effective. 

\item 
  As listed in Table \ref{tab:daibunrui}, 
  when determining ``Other'' tenses/aspects/modalities (those other than 
  ``Present'' and ``Past''), 
  Method 2, using the result of language analysis, 
  yields higher accuracy rates than those of Method 1 or the software. 
  
  It is important to examine 
  the accuracy rate when determining difficult tenses/aspects/modalities 
  if we want to implement 
  high-quality machine translation. 
  Even if the overall accuracy rate (``All'') is high, 
  high-quality translations will not be produced 
  if only ``Present'' or ``Past'' are selected correctly. 
  Although the overall accuracy of Method 2 is 
  a little low,   the accuracy for determining the difficult tenses/aspects/modalities 
  ``Other'' is high. 
  We therefore think that 
  Method 2 is more promising for high-quality machine translation 
  than Method 1. 

\end{enumerate}

\item
Problems of our method 
\begin{enumerate}
\item 
In Japanese, 
two sentences that 
have the same surface expression for the verb phrase 
sometimes 
have different tenses/aspects/modalities. 
For example, in the two-verb phrases 
{\it tokeru} ``can solve'' or ``thaw'' in the following examples, 
the first one has the modality ``Potential'' 
and the second one has only 
the tense ``Present.'' 

{\small
\vspace{0.2cm}
  \begin{tabular}[h]{llll}
shougakusei-nara & taiteiwa & konomondai-wa & \underline{tokeru.}\\
(elementary schoolchildren) & (most) & (this problem) & (can solve)\\
\multicolumn{4}{l}{Most elementary schoolchildren \underline{can} solve this problem.} \\
  \end{tabular}

\vspace{0.2cm}
  \begin{tabular}[h]{llll}
ike-no & koori-wa & sangatsu-ni & \underline{tokeru.}\\
(pond) & (ice)   & (in March) &  (thaw)\\
\multicolumn{4}{l}{The pond \underline{thaws} in March.} \\
  \end{tabular}
\vspace{0.2cm}
}

To handle these examples, 
we must disambiguate the word senses of 
{\it tokeru}: ``can solve'' or ``thaw.'' 

\item 

Although our method uses
only Japanese tense/aspect/modality expressions 
and does not consider 
the structure of the English translated sentence, 
the tense/aspect/modality that should be used sometimes depends 
on the structure of the translated English sentence. 
The first of the following sentences 
has the aspect ``Non-Progressive,'' and the second has 
the aspect ``Progressive.'' 

{\small
\vspace{0.2cm}
  \begin{tabular}[h]{llll}
     kare-wa & shitsujituna & seikatsu-wo & \underline{okut-teiru.}\\
 (He) & (sober and simple) & (life) & (live)\\
  \multicolumn{4}{l}{He \underline{lives} a sober and simple life.}\\
  \end{tabular}

\vspace{0.2cm}
\begin{tabular}[h]{llll}
kare-wa & taidana & seikatsu-wo & \underline{okut-teiru.}\\
 (He) & (lazy) & (life) & (be leading)\\
 \multicolumn{4}{l}{He \underline{is leading} a lazy life.}  \\
  \end{tabular}
\vspace{0.2cm}
}

We can consider that 
the verbs of these Japanese sentences have almost the same meaning, 
and the same tense/aspect/modality. 
But changing the verb used in English translation 
from ``live'' to ``lead'' makes 
the difference between ``Present'' and ``Progressive.'' 
If we want to use our method in high-quality translation, 
it is necessary to match not only Japanese sentences 
but also English sentences  
when detecting a similar example. 
In an overall machine translation system, 
the structure of the English translation of 
a Japanese input sentence is made up of results 
of the structure analysis. 
By using the results, 
we will only be able to detect examples whose structure is similar to 
the structure of the English translation. 

\item 
In some cases 
it would be better to use 
not only expressions at the end of the sentence 
but also adverbs at the beginning \cite{kume90e}. 
For example, 
the difference between \citeform[already]{mou} and 
\citeform[yesterday]{kinou} makes 
the difference between ``Past perfect'' and ``Past.'' 

{\small  
\vspace{0.2cm}
  \begin{tabular}[h]{lll}
\it \underline{mou} & \it touroku-shita.  & I\underline{'ve} already registered.\\
 (already)      & (register)  & \\[0.3cm]
\it \underline{kinou}& \it touroku-shita. & I register\underline{ed} yesterday.\\
 (yesterday)      & (register)  & \\
  \end{tabular}

\vspace{0.2cm}
}

\None{
Although 
we may be able to handle such a case 
by increasing the number of examples and 
matching from the end to the beginning, 
it is thoughtless to expect to increase 
the number of examples. }
Our method would have to be 
changed if it were to 
handle the above case. 

\end{enumerate}

\None{
\item
Notice of our experiments 
%
  In the experiment 
  the criterion for judging 
  whether the result was correct was 
  very strict: 
  the output modality is 
  the same as the modality of the English translation 
  in our bilingual database of the input sentence. 
  Because we originally assume various English translations 
  against one Japanese sentence, 
  it is not necessary to 
  translate to the same as the English translation that 
  a Japanese input sentence has. 
  A bilingual person who is very good at 
  both Japanese and English and who 
  checks the experimental results, 
  might think almost all of them are correct. 
  Looking at Table \ref{tab:kasanari}, 
  you will notice that 
  there are many cases in which both our method 
  and the software gave incorrect answers. 
  This indicates that 
  although both our method and the software 
  give correct answers, 
  the English translation that an input sentence has 
  is eccentric, and 
  the results of both methods are judged to be incorrect. 
}

\item
Advantages of our method 

\begin{enumerate}
\item 
  It does not require hand-craft rules. 
\item 
  It is very easy to implement. 
\end{enumerate}
Our method determined tense/aspect/modality 
more accurately 
than the top-level MT software currently available  on the market. 
This indicates that 
our method is useful. 
\end{enumerate}

\section{Conclusion}

To translate Japanese tense/aspect/modality expressions 
into English by using the example-based method, 
we defined the similarity between input and example sentences  
as the degree of semantic match 
between expressions at the end of sentences. 
We used the k-nearest neighbor method 
in order to exclude the effects of noise. 
In experiments, our method translated tense/aspect/modality 
expressions more accurately than 
the top-level MT software currently available on the market did. 
Another advantage of our method is that 
it does not require hand-craft rules. 
\None{
The two reasons of the higher accuracy rate and 
the needless of manually customization indicate 
that our method is very excellent. }

We used two methods to evaluate the degree of 
similarity: one that simply matches character strings, 
and the other that uses the result of language analysis. 
The overall accuracies obtained by using the string-matching are only 
a little better than those obtained by using 
language analysis. However, the results of translating 
tenses/aspects/ modalities 
other than ``Present'' and ``Past'' are quite a bit better 
when the language analysis was used. 
Because high-quality machine translation requires 
effective handling of difficult tenses/aspects/modalities, 
we think that the latter method will be more promising. 

The tense/aspect/modality translation method we developed can also 
be applied to 
English-to-Japanese translation 
by eliminating the subject of the English input sentence
and using string-matching from the beginning of the remainders; 
that is, from the beginning of a verb phrase. 
\None{Since this method does not need the technical knowledge, 
performing this work by people who are not Japanese is easy. }
And because this method does not need hand-craft rules, 
it is very useful for many other languages 
where hand-craft rules have not been prepared well. 
We will also be able to use our method 
for monolingual tense/aspect/modality analysis. 
For example, 
if instead of the bilingual corpora 
we use the monolingual corpora 
tagged with the correct tense/aspect/modality, 
we will be able to identify the tense/aspect/modality immediately.

\bibliographystyle{tmi}

\end{document}